\documentclass[a4paper, runningheads]{llncs}

\usepackage{makeidx}  
\usepackage{amssymb}
\usepackage{amsmath}
\usepackage{listings}
\usepackage{xcolor}
\usepackage{url}
\usepackage{mathtools}

\begin{document}
\mainmatter              
\title{Declarative Guideline Conformance Checking of Clinical Treatments: A Case Study}
\titlerunning{Declarative Guideline Conformance Checking of Clinical Treatments}  
%
\author{Joscha Grüger\inst{1}\inst{2} \and Tobias Geyer\inst{2} \and Martin Kuhn\inst{2} \and Stephan A. Braun\inst{3}\inst{4} \and Ralph Bergmann\inst{1}\inst{2}}
\authorrunning{Joscha Grüger et al.}   
%
\tocauthor{Joscha Grüger, Tobias Geyer, Martin Kuhn, Ralph Bergmann}

\institute{University of Trier, Business Information Systems II, Trier, Germany,\\
\email{grueger@uni-trier.de}
\and
German Research Center for Artificial Intelligence (DFKI),
SDS Branch Trier, Trier, Germany,\\
\and
Dept. of Dermatology, University Hospital M\unexpanded{\"u}nster, M\unexpanded{\"u}nster, Germany\\ \and
Dept. of Dermatology, Medical Faculty, Heinrich Heine University, Düsseldorf, Germany
}

\maketitle              

\begin{abstract}        
Conformance checking is a process mining technique that allows verifying the conformance of process instances to a given model. Thus, this technique is predestined to be used in the medical context for the comparison of treatment cases with clinical guidelines. However, medical processes are highly variable, highly dynamic, and complex. This makes the use of imperative conformance checking approaches in the medical domain difficult. Studies show that declarative approaches can better address these characteristics. However, none of the approaches has yet gained practical acceptance. Another challenge are alignments, which usually do not add any value from a medical point of view. For this reason, we investigate in a case study the usability of the HL7 standard Arden Syntax for declarative, rule-based conformance checking and the use of manually modeled alignments. Using the approach, it was possible to check the conformance of treatment cases and create medically meaningful alignments for large parts of a medical guideline.

\end{abstract}
\section{Introduction} 
Clinical guidelines are systematically developed statements that reflect the state of evidence-based medical knowledge at the time the guidelines are created. They are intended to support physicians and patients in the decision-making process for appropriate medical care in specific clinical situations \cite{Lohr.1990}. Thus, they represent the documentation of evidence-based medicine and are an important tool for the scientifically sound treatment of patients. However, it is not yet possible to say to what extent guideline knowledge is applied in treatment \cite{Forsner.2010,Landfeldt.2015}. Although some approaches were presented, no holistic standard for the representation and verification of guideline compliance in clinical treatment has yet become established. The investigation of guideline compliance or non-compliance is of medical interest though and holds potential insights for research. For this purpose, the process perspective can provide essential information, e.g., to research observed behavior such as differences between the executed activities and the activities recommended in the guidelines \cite{Rovani.2015} and to help to understand the corresponding reasons and implications, which is in the interest of physicians \cite{Gatta.2019}.


The detection of deviations and the evaluation of the conformity of process instances is part of process mining. Process mining is an emerging field of research and fills the gap between data mining and business process management \cite{vanderAalst.2016}. One technique of process mining is conformance checking, whose approaches focus on measuring the conformance of a process instance to a process model. The results of the measurement can usually be output in the form of metrics or alignments \cite{Adriansyah.2012}, i.e., corrective adjustments for process instances.


Previous research attempted to convert guideline knowledge into imperative representations of a data Petri net and use this to check conformance \cite{Geleijnse.2018,Gruger.2022}. The result is that the high degree of variability and flexibility of medical processes makes the complete imperative modeling of all variants almost impossible \cite{Gruger.2022}. 

One way to address the stated difficulties is to use declarative process models. Declarative process modeling languages are considered to be particularly suitable for representing processes in the medical context \cite{Bottrighi.2009,Burattin.2016}. Unlike imperative approaches, declarative modeling allows the definition of rules for the execution of processes without excluding non-modeled process behavior \cite{Pesic.2006}. This reflects well the nature of medical processes and guidelines and takes into account the unpredictability of events over the course of treatment.

Both the imperative and declarative conformance checking approaches provide alignments that are technically appropriate. However, they are not meaningful from a medical perspective. From the perspective of domain experts, these alignments include adjustment instructions that are inappropriate for the use and evaluation of the results. For example, it does not make sense to delete patients' diseases or change the guideline as part of the alignment simply because this can optimize the costs of the alignment \cite{Gruger.2022}.

In this case study, we investigate to which extent the mentioned problems of imperative process models trying to cover all contingencies and, more generally, the medical meaningfulness of alignments can be addressed in a declarative approach. For this purpose, we use the Arden Syntax for Medical Logic Modules (MLMs), which is widely used in medicine and HL7 standardized, to formalize medical knowledge \cite{Samwald.2012}. Besides the classical use of Arden syntax, we show how it can be used as a declarative description of process models by defining rules. The formalism enables rule-based declarative representation of guideline knowledge. So far, the standard Arden Syntax has found application in decision support \cite{Anand.2018,Schuh.2018,Adlassnig.2015}. Despite the expressive power of the syntax, there are no approaches to conformance checking. The challenge of meaningful alignments is addressed via manually developed alignment steps that are integrated in the respective MLM. The presented approach is applied and evaluated on real patient data.

The remainder of the paper is organized as follows. Section 2 provides background information on the components of our approach. Section 3 describes the research method for the model generation, the conformance checking approach as well as the alignments and shows how the event log from the patient data is created. Section 4 presents the implementation and the results. In Section 5, the findings are discussed and Section 6 concludes the paper.

\section{Fundamentals}\label{sec:fundamentals}

\subsection{Conformance Checking and Alignments}
\textit{Conformance checking} describes the process of identifying discrepancies between the desired behavior of the process depicted by the process model and the real behavior of the process depicted by the event log \cite{Rovani.2015}. Standard conformance checking only considers the control-flow perspective \cite{Caron.2013} but event logs often contain information, that go beyond the ordering of events such as time or data related information \cite{Caron.2013}. Since multiple perspectives can be considered, deviations are not only possible in the ordering of the events but also in other perspectives. 
One type of conformance checking is \textit{local conformance checking}. This type describes the process of checking the conformance by using a set of independent rules regarding the process. Therefore, only specific parts of the process are checked and not the process as a whole. These rules are often defined in linear temporal logic or in declarative modeling languages. \cite{Caron.2013}  

Most state-of-the-art conformance checking techniques are using \textit{alignments} \cite{Dunzer.2019}. An alignment can be seen as mapping of the process model capturing the desired behavior and the event log recording the behavior that occurs in reality \cite{vanderAalst.2016}. Also, the concept of alignments is not process modeling language specific and thus can be defined for nearly all modeling languages \cite{vanderAalst.2016}. 

For the creation of an alignment, the events recorded in the event log are mapped to process steps in the process model. Alignments are defined by moves. A \textit{log move} is executed when an event is recorded in the event log that does not occur in the model. A \textit{model move} is executed when an event that is required by the process model is missing in the event log. If the mapping is correct from a control flow perspective but violates a condition from a data perspective, an \textit{incorrect synchronous move} is executed. If the control flow and data perspective match, it is called a \textit{synchronous move}. Using these alignment moves, it is possible to identify the events that violate the process execution \cite{Adriansyah.2013,Leoni.2013,MannhardtDiss.2018}.

\subsection{Medical Logic Modules and Arden Syntax}
Medical Logic Modules (MLMs) were developed with the aim of presenting medical knowledge in self-contained units, readable by humans and interpretable by computers, and transferable to other clinics \cite{Arkad.1991,Hripcsak.1994}. The Arden Syntax is a declarative, HL7-standardized, open implementation of MLMs \cite{Samwald.2012,Vetterlein.2010}. It was developed for the exchange of medical knowledge. In the following, we interpret the term MLM as MLM in the Arden Syntax. MLMs are text files arranged in discrete slots. Each slot has a name (e.g., \texttt{version}) followed by a colon and the body (e.g., \texttt{2.0}) and ends with a semicolon. Depending on the slot, the body can contain free text, code, or structured data. To improve readability, MLMs are divided into three categories: \textit{maintenance}, \textit{library}, and \textit{knowledge} \cite{Hripcsak.1994}.

Maintenance contains slots not directly related to medical knowledge, rather information such as title, author, and version. The library category contains slots related to the source of the modeled knowledge, keywords and textual explanations. Knowledge category slots describe what the module does. It is divided into three sections. First, the \textit{data} slot, which can be used to retrieve patient data from a database. Second, the \textit{evoke} slot, which specifies conditions when a module is activated. Last, the \textit{logic} slot contains the actual medical rule or medical condition to be evaluated  \cite{Arkad.1991}. Therefore, the Arden Syntax provides extensive operators. Such operators as \texttt{before}, \texttt{after}, \texttt{within same day} or \texttt{n days before/after} natively allow addressing the time perspective \cite{ImplementationGuides}. \textit{Action} slot defines what to do when the logic slot is evaluated against \texttt{true} \cite{Arkad.1991}. For example, an MLM could be evoked when a new patient is admitted. In the logic slot, it could be checked whether the patient has a food intolerance. If the condition was evaluated as \texttt{true}, a request for nutritional counseling could be entered into the hospital information system (HIS) database.

\section{Materials and Methods} \label{sec:ResearchMethod}
\begin{figure}
    \centering
    \includegraphics[width=1.0\textwidth]{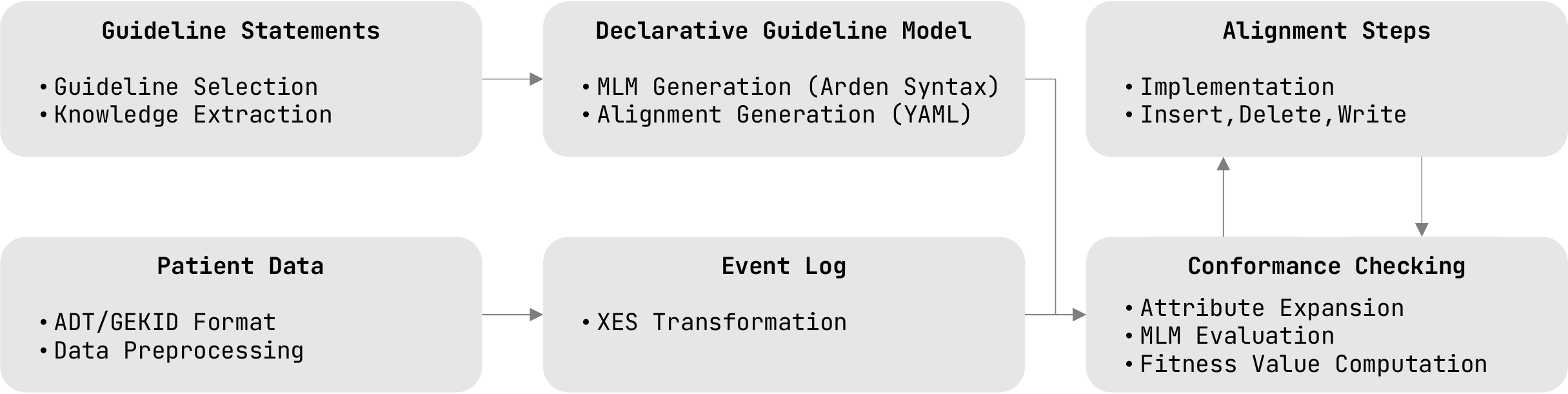}
    \caption{Methodology Outline.}
    \label{fig:methodology}
\end{figure}

The methodology is outlined in Figure \ref{fig:methodology}. In the first step, the guideline knowledge to be used is selected and then transformed into the declarative model. The focus is on the complete and correct transformation of the guidelines' medical knowledge formulated in free text into the declarative, computer-interpretable representation. In parallel, the event log is created from the patient data. Finally, conformance checking is performed where alignments are applied until the trace is conforming to the model, and the degree of guideline compliance can be evaluated based on the results.

\subsection{Generation of the MLM Rule Base}
The declarative approach of MLMs and Arden Syntax is used to represent the guideline knowledge as a declarative rule-based model for conformance checking. For this purpose, an MLM is created for each of the guideline statements. Thus, the guideline is represented by a set of MLMs. MLMs offer the advantage that, despite their expressive power, they are readable in parts even for domain experts after a short training  \cite{Samwald.2012}.  In the context of clinical guidelines, the expressiveness results from the already existing fields of the MLM structure covering the majority of the information fields of a guideline statement, the extensibility of these by new data fields, and the usability of the Arden Syntax and SQLite.

In principle, MLMs are designed to be evoked by an external call, an event or an evoke condition (\textit{evoke slot}). During an evocation the logical part is evaluated (\textit{logic slot}) and depending on the result an action can be executed (\textit{action slot}) \cite{Samwald.2012}. The approach of this case study uses the expressive power of MLMs to formulate rules for detecting deviations based on the events of the log. Therefore, the evoke slot is used to define and subscribe to events of the event log that trigger the respective MLM. Since the data perspective also can trigger MLMs in addition to the control flow perspective, data attributes are also interpreted as independent events, so-called \textit{attribute writes}, in the evocation process. In the present conformance checking approach, the use of logic and action slots is redefined to check the compliance of treatment cases to a given guideline statement and return the appropriate alignment. For this purpose, the entire conformance with the guideline statement is checked in the logic slot and the result is returned. A distinction is made between conforming and non-conforming traces. Conforming traces only return the describing metadata for the examined statement. Non-conforming traces return previously by domain experts defined alignment steps (see \ref{sec:alignments}). In the action slot, the conformance is returned as a boolean value together with the alignment generated in the logic slot.


The process of transforming clinical guideline knowledge into explicit, com\-pu\-ter-interpretable formalizations is complex and fraught with responsibility. Misinterpretations can have serious consequences for patient health \cite{Patel.1998}. For correct interpretation of guideline statements, implicit background knowledge of different fields of expertise is required. To ensure the correct transformation of guideline statements into MLMs, the CGK4PM framework was applied \cite{Gruger.2022b}. Following this, an interdisciplinary team of doctors and computer scientists was assembled. These worked together in the iterative approach of formalization and validation described in CGK4PM. Formalization took place in workshops, in which 5 statements were discussed in each session. Validation was done firstly by the doctors in workshops and secondly technically, with test patients representing compliant and non-compliant patients. After the joint formalization workshops, the cases were transferred to the Arden Syntax by the computer scientists and are validated using unit tests.

Exemplary, a part of the German clinical guideline for the treatment of malignant melanoma of the Association of the Scientific Medical Societies (AWMF)\footnote{\url{https://www.awmf.org}} was transferred into MLMs. The focus was on Chapter 4 (Diagnostics and Therapy in Primary Care) and Chapter 6 (Diagnostics and Therapy in Locoregional Metastasis). For this purpose, 39 guideline statements were randomly selected and each statement was transferred into an MLM. Five of these statements were optional and therefore not relevant for conformance and not transformed. Nine statements were not transitioned due to lack of data in the underlying database, like structured information about uncertainties and suspicions.

The model results in 25 individual and independent MLMs and thus represents as many guideline statements. In the knowledge category, the data, evoke, logic and action slots are used for the actual evaluations. In the data slot, all data needed to check compliance is retrieved from the database and provided to the evoke, logic, and action slots for evaluation. Moreover, the events relevant for MLM evocation are defined here. The data query step turned out to be the most difficult and error-prone step, since at this point the information required for the logic part must be specifically queried using SQL syntax and Arden Syntax. 

\subsection{Event Log}\label{sec:eventlog}
To maintain comparability, the same dataset as in \cite{Gruger.2022} was used. The dataset contains data on five patients with malignant melanoma and their treatment at Münster University Hospital. Patients were informed about the use of data and gave their written consent. For privacy reasons, the data were made available to the research in anonymized form. The data are in the uniform XML format of cancer registries in Germany, ADT/GEKID\footnote{https://www.gekid.de/adt-gekid-basisdatensatz}. Among other things, it contains data on treatment, diagnosis, histology, and cancer staging. The major advantage of using data in the format of the basic data set is that it is used by all German cancer registries and results are thus transferable and comparable. In addition, all entries contain a timestamp, which enables procedural use. 

For conformance checking, the data is converted to the XES event log format \cite{XES}. For this purpose, a generic XML to XES converter was implemented in Python and configured to convert ADT/GEKID data to XES. During the preprocessing in \cite{Gruger.2022}, an extensive filtering of activities according to relevance and irrelevance for guideline compliance (e.g., psychological counseling) was necessary because conformance checking with an imperative model was used. Contrarily, in the present approach, conformance checking is done on the whole log without filtering since the declarative method by itself considers only relevant data.
However, the naming of attributes and events in the event log had to be aligned with the model.

\subsection{Conformance Checking}
MLMs subscribe to events, these can relate to both control flow level events and data level write operations. Therefore, the trace is extended so that attribute writes are also interpreted as events for evocation and thus conformance checking is also evoked for each write. It is important to note that the MLMs subscribe to the writing of a specific attribute value, e.g., for the attribute \texttt{ICD-Code} with the value \texttt{C43.9} the event \texttt{write\_icd\_code\_c43.9} would be triggered. 

In the evoked MLMs, conformity is first calculated locally and expressed as \textit{conform} (1) or \textit{non-conform} (0). Then, a global computation of a fitness value is performed across all MLMs. It should be noted that not every MLM is evoked for every trace. Therefore, the calculation is performed including only the evoked MLMs. With respect to the fitness dimension of conformance checking, we quantify the conformance of a trace $\sigma$ from Log $L$ with an MLM model $M$. Therefore, we introduce the fitness function $fitness(\sigma,M)$. This returns a value between 0 and 1, quantifying the degree of conformance of trace $\sigma$ with the model $M$. Here, 1 represents optimal fitness and 0 represents very poor fitness.

\begin{definition}[Fitness]
Let $MLM$ be the universe of all MLMs, $M\subseteq MLM$ be the declarative model, and $\sigma$ be a trace. Then $M_{\sigma}\subseteq M$ is the short form of all MLMs evoked for sigma. The function $eval:M\times \Sigma \to \{0,1\}$ evaluates whether a trace conforms to an MLM or not or was not evoked. The fitness is defined as:
\[
   fitness(\sigma, M) = \frac{\sum_{i=1}^{n} eval(\sigma,M'_i)}{|M'_{\sigma}|}
\]
\end{definition}

\subsection{Alignment}\label{sec:alignments}
In \cite{Gruger.2022}, it is shown that from both a medical and guideline compliance use case perspective, it is not reasonable to align the model. Therefore, the alignment is done under the assumption that the model is correct and only the trace is adjusted. Thus, a log move is always handled with a \texttt{DELETE} operation in the event log, a model move with an \texttt{INSERT} in the event log, and an incorrect synchronous move is handled with a data \texttt{WRITE} in the event log. For this purpose, alignment steps have been manually defined for each MLM. 
Each alignment step consists of a set of $m$ alignment operations. Each operation consists of an operator, a positional relation, a value, at least one timestamp as position information and optional parameters concerning the operation. 
\begin{definition}[Alignment Steps]
Let $O=\{INSERT, DELETE, \\WRITE\}$ be the set of all alignment operations, $R=\{BEFORE, AFTER,\\ AT, BETWEEN\}$ be the set of positional relations, $S=\{ATTRIBUTE, \\EVENT\}$ be the set of the subjects involved, $V$ be the universe of all possible attribute values, and $TS$ be the universe of all possible timestamps. Then an alignment proposition is a quadruple $(s,o,v,p)$ with $s\in S, o\in O, v\in V$ and $p \in R\times TS^*$.
\end{definition}

In the MLMs, the alignment steps are YAML encoded. The alignment engine (AE) then resolves the received alignment steps and performs the operations described in them. For this purpose, based on the position information in the form of one or two timestamps and the positional relation for each \texttt{INSERT}, a matching position is searched for in the trace. Either the earliest conforming position, the last position or a random position can be selected. In case of a \texttt{DELETE} or \texttt{WRITE}, the exact event must be identified by timestamp and activity name. If no conforming position can be found, the alignment is aborted. If the alignment was successful, the trace is run through again to check the effect of the alignment on previous events and their conformity. 

\begin{figure}
    \centering
    \includegraphics[width=1.0\textwidth]{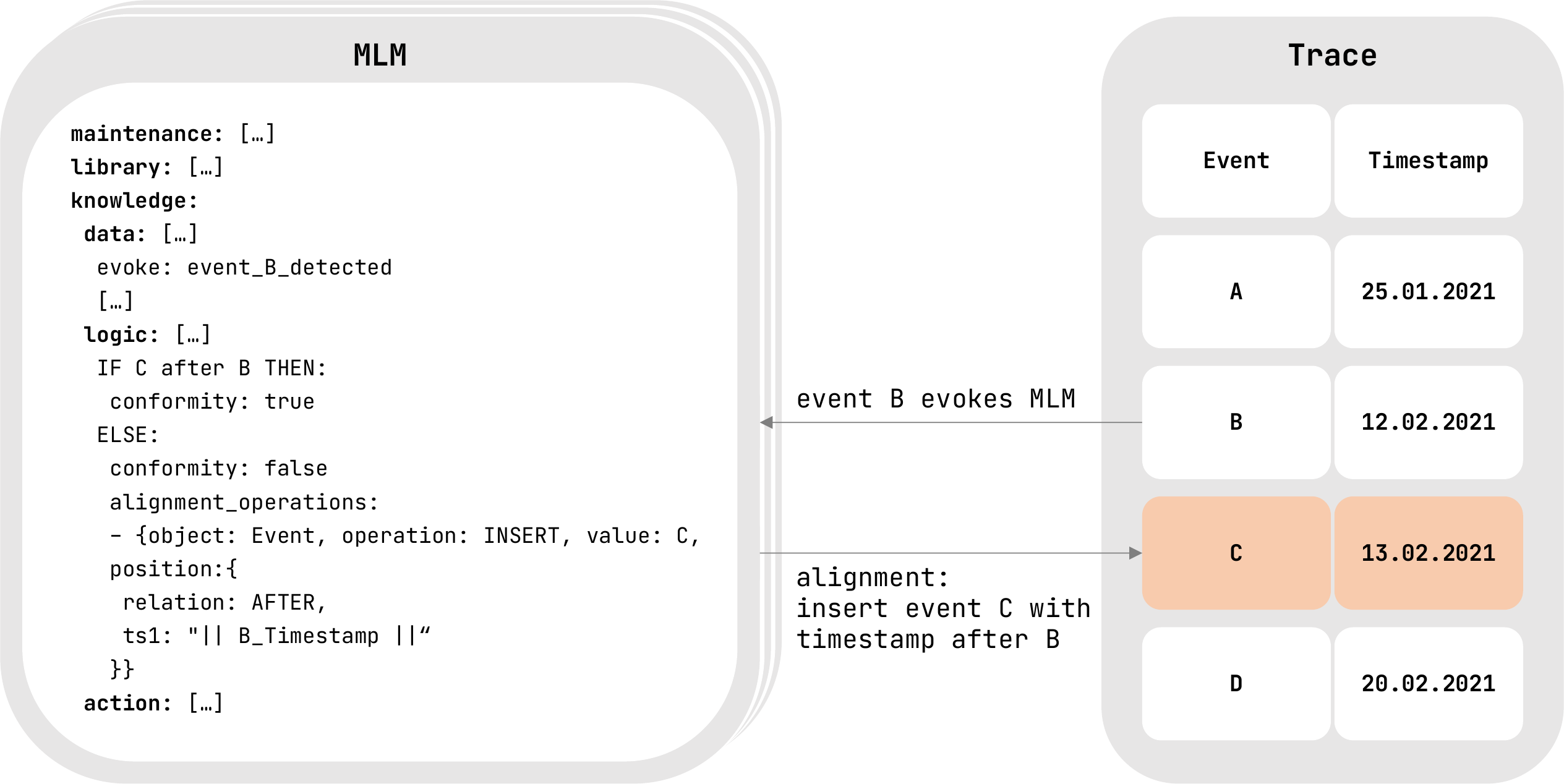}
    \caption{The outlined example shows a trace that violates the MLM, which states that event \texttt{B} must be followed by event \texttt{C}. The alignment step modeled manually in the MLM indicate that \texttt{C} is to be inserted after \texttt{B}.}
    \label{fig:examplealignment}
\end{figure}

Figure \ref{fig:examplealignment} shows an outlined example of an alignment calculated with an MLM. In this MLM, event \texttt{C} should occur after event \texttt{B}. The MLM is evoked since event B occurs in the trace. Since event \texttt{C} does not occur, the logic of the MLM concludes to \texttt{false} and the alignment operation defined in the \texttt{else} block is executed. Thus, an event is generated with the value of \texttt{C} and is inserted after the \texttt{B} event. To find the correct position for inserting, the timestamps of the events are used. The end result is the insertion of event \texttt{C} in the trace and the detection of the guideline violation due to the initial absence of \texttt{C}.

\section{Implementation}
\begin{figure}
    \centering
    \includegraphics[width=0.9\textwidth]{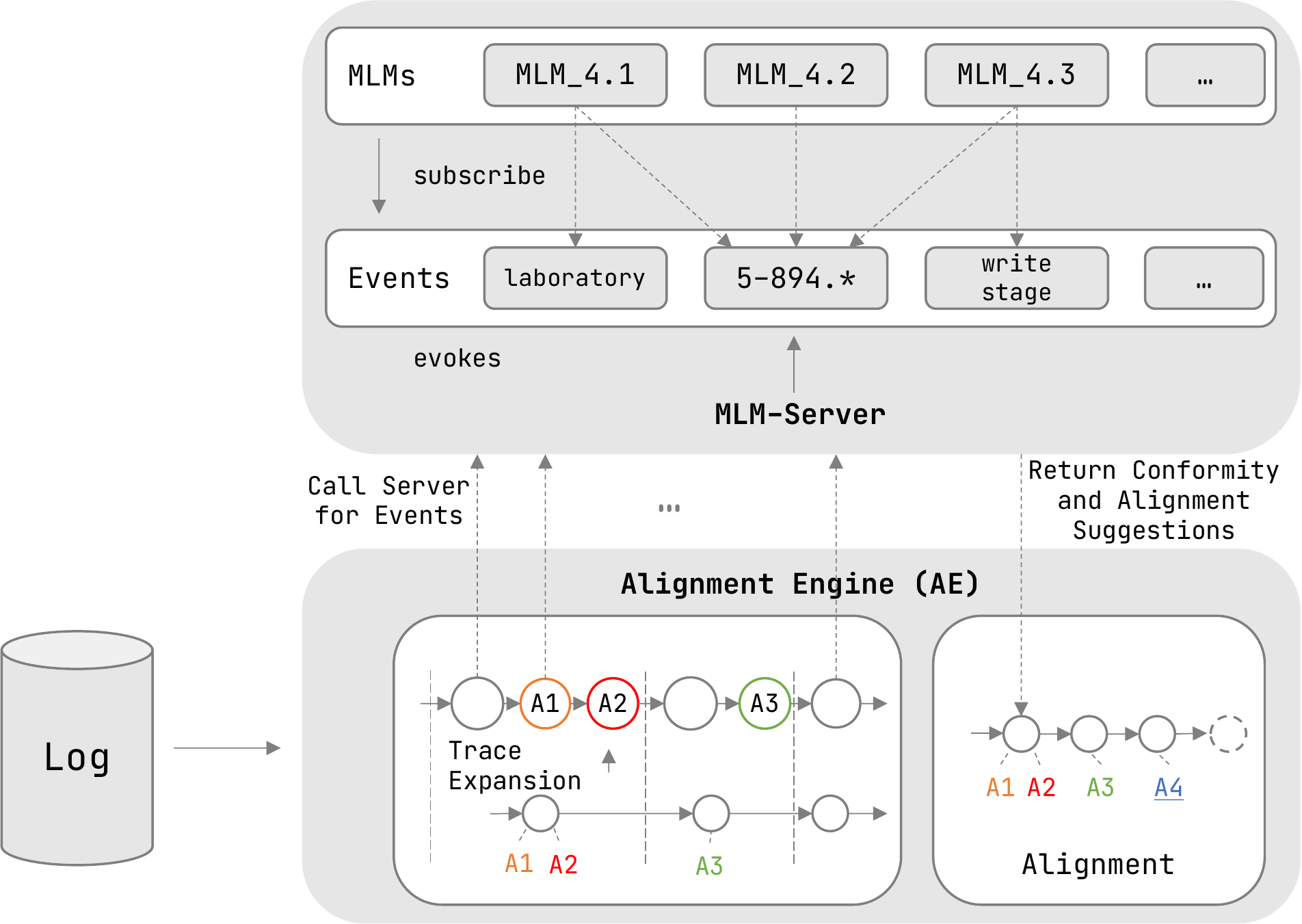}
    \caption{The AE expands the traces from the log and generates an event for each write operation of an attribute. For each event, the MLM server is called, which then evokes MLMs subscribed to the respective event. Based on the returned manually modeled alignment steps, the AE generates a new compliant trace without changing the model.}
    \label{fig:systemstructure}
\end{figure}

The implementation was split into the central management of the rule-based model and events in a server component (\textit{MLM server}) and the local implementation of the rules in the client (\textit{Alignment Engine (AE)}). The interaction of the components is depicted in Figure \ref{fig:systemstructure}.
In essence, the MLM server triggers the MLMs relevant to an event and returns the results to the AE. For this purpose, the MLM server manages subscriptions of events by the MLMs, receives traces and events from the AE, triggers the MLMs subscribing to the received event, summarizes the results of all evaluations and returns them to the AE. The server component was implemented in Python, using the Arden2ByteCode compiler.

The AE expands traces so that attributes are transformed to attribute write events. Sequentially, the AE calls the MLM server for each of the events. For this purpose, the trace is transferred to a temporary SQLite database. In case of an MLM violation, the alignment steps provided by the server are converted into YAML. To do this, the AE resolves the positional relations to the timestamps given in the alignment steps and thus identifies the possible position(s) in the trace to implement the alignment. The alignment process is restarted with the aligned trace to consider also implications that have arisen through the implementation of the alignment.
 
The approach was evaluated based on five real world treatment cases (see \ref{sec:eventlog}). Each trace consists of 29 events on average, and each event has eight attributes on average. In total, counting only the first evocations per MLM per trace, 47 evocations were performed. Of these, 21 checks failed and 26 were compliant. All alignments were successfully applied by the AE. A total of 21 alignments were executed, 15 deletes and six inserts. No verification failed on the four MLMs that write attributes based on alignments. On average, the traces have a fitness of 0.478, while the worst fitness is 0.2 and the best is 0.66.

\section{Discussion} 
Considering the results, significant advantages could be identified in the use of the declarative modeling approach. In a qualitative comparison with the imperative approach from \cite{Gruger.2022}, working on the same guideline and same data set, most limitations mentioned there could be addressed with the declarative approach. However, mapping ambiguities remain a challenge \cite{Gruger.2022}. This means that it is not always possible to determine whether the event was executed because of a particular attribute, or whether that attribute is just an unnoticed byproduct. The Arden Syntax natively allows inclusion of the time perspective and implementation of softer temporal constraints like delays. The declarative nature of the approach made it possible to handle high variability of processes. In addition, the semantic meaningfulness of the alignments could be improved by pre-modeled steps. Limitations still exist in the alignment of events, especially in processes running in parallel in one instance, e.g., the treatment of comorbidities during a hospital stay. Furthermore, explicit mapping of repetitive activities to activities in the model remains a challenge. In addition, the fitness score is still not very meaningful, as it does not take into account the severity of the deviation.

The transformation of the guideline to the rule-based model proved to be complex. This is due to the specific knowledge to be modeled and to the lack of verification tools that, e.g., search for contradictions or interdependencies in the rule base. The advantage of modeling with Arden Syntax was that after some practice, the domain experts were also able to read and edit parts of the MLMs.

The use of manually modeled alignment steps required a high initial effort. From the user's perspective, however, these addressed the deviations semantically correct contrary to other algorithmic approaches, e.g., the one used in \cite{Gruger.2022}. However, one limitation related to alignments is the runtime of the alignment implementation in the trace. Since each alignment can also affect previous events, the entire trace must be checked again for conformance.
\section{Conclusion}
In this paper, we focused on the MLM-based declarative modeling of guidelines and a conformance checking approach that meaningfully considers guideline characteristics and medical context. The declarative modeling approach excelled in its expressiveness and its characteristic of dealing with the variability of medical processes. The case study showed promising initial results by considering medically meaningful logic and individually defined alignment steps. These novel features are the key difference from conventional conformance checking algorithms and contribute to medical application of declarative process mining. However, there are still severe limitations in the scope of the expressions (e.g. before/after), which only refer to the timestamp and not to the activities.

We plan to add more guideline statements to the model for further evaluation and identification of further challenges. Moreover, it should be investigated to what extent ArdenML (an XML schema for expressing MLMs) and its possibility to connect to other XLM-based HL7 standards can be beneficially integrated into the architecture. Furthermore, we want to improve the conformance checking approach and tackle the discussed limitations such as parallel and repetitive activities. Since the algorithm can currently only evaluate if an activity is compliant or not, we plan to extend it with a fuzzy function. Thus, we hope to make more meaningful statements about the treatments by differentiating more precisely between guideline violations such as temporal deviations. These details that distinguish the patients with regard to treatment outcomes can be highly relevant for physicians in their reasoning process.

\section*{Acknowledgments}
The research is partially funded by the German Federal Ministry of Education and Research (BMBF) in the project DaTreFo under the funding code 16KIS1644.
%
%

\bibliography{references}

\begin{thebibliography}{10}

\bibitem{Lohr.1990}
Lohr, K.N., Field, M.J.:
\newblock {Clinical practice guidelines: Directions for a new program}. Volume
  90-08 of {Publication IOM}.
\newblock {National Academy Press}, Washington (1990)

\bibitem{Forsner.2010}
Forsner, T., Hansson, J., Brommels, M., Wistedt, A.A., Forsell, Y.:
\newblock Implementing clinical guidelines in psychiatry: a qualitative study
  of perceived facilitators and barriers.
\newblock BMC psychiatry \textbf{10} (2010) ~8

\bibitem{Landfeldt.2015}
Landfeldt, E., Lindgren, P., Bell, C.F., Schmitt, C., Guglieri, M., Straub, V.,
  Lochm{\"u}ller, H., Bushby, K.:
\newblock Compliance to care guidelines for duchenne muscular dystrophy.
\newblock Journal of neuromuscular diseases \textbf{2}(1) (2015)  63--72

\bibitem{Rovani.2015}
Rovani, M., Maggi, F.M., {de Leoni}, M., {van der Aalst}, W.M.:
\newblock Declarative process mining in healthcare.
\newblock Expert Systems with Applications \textbf{42}(23) (2015)  9236--9251

\bibitem{Gatta.2019}
Gatta, R., Vallati, M., Fernandez-Llatas, C., Martinez-Millana, A., Orini, S.,
  Sacchi, L., Lenkowicz, J., Marcos, M., Munoz-Gama, J., Cuendet, M., de~Bari,
  B., Marco-Ruiz, L., Stefanini, A., Castellano, M.:
\newblock Clinical guidelines: A crossroad of many research areas. challenges
  and opportunities in process mining for healthcare.
\newblock In Di~Francescomarino, C., Dijkman, R., Zdun, U., eds.: Business
  Process Management Workshops, Cham, Springer International Publishing (2019)
  545--556

\bibitem{vanderAalst.2016}
van~der Aalst, W.M.P.:
\newblock Process Mining: Data Science in Action. 2nd edn.
\newblock {Springer Berlin Heidelberg} (2016)

\bibitem{Adriansyah.2012}
Van~der Aalst, W., Adriansyah, A., van Dongen, B.:
\newblock Replaying history on process models for conformance checking and
  performance analysis.
\newblock Wiley Interdisciplinary Reviews: Data Mining and Knowledge Discovery
  \textbf{2}(2) (2012)  182--192

\bibitem{Geleijnse.2018}
Geleijnse, G., Aklecha, H., Vroling, M., Verhoeven, R., Van~Erning, F.,
  Vissers, P., Buijs, J., Verbeek, X.:
\newblock Using process mining to evaluate colon cancer guideline adherence
  with cancer registry data: a case study.
\newblock In: AMIA 2018, American Medical Informatics Association Annual
  Symposium: San Francisco, California, USA, 3-7 November 2018.
\newblock American Medical Informatics Association (2018)

\bibitem{Gruger.2022}
Gr{\"u}ger, J., Geyer, T., Kuhn, M., Braun, S.A., Bergmann, R.:
\newblock Verifying guideline compliance in clinical treatment using
  multi-perspective conformance checking: A case study.
\newblock In Munoz-Gama, J., Lu, X., eds.: Process Mining Workshops. Volume
  433.
\newblock {Springer International Publishing} and {Imprint Springer}, Cham
  (2022)  301--313

\bibitem{Bottrighi.2009}
Bottrighi, A., Chesani, F., Mello, P., Molino, G., Montali, M., Montani, S.,
  Storari, S., Terenziani, P., Torchio, M.:
\newblock A hybrid approach to clinical guideline and to basic medical
  knowledge conformance.
\newblock In Combi, C., Shahar, Y., Abu-Hanna, A., eds.: Artificial
  Intelligence in Medicine, Springer Berlin Heidelberg (2009)  91--95

\bibitem{Burattin.2016}
Burattin, A., Maggi, F.M., Sperduti, A.:
\newblock Conformance checking based on multi-perspective declarative process
  models.
\newblock Expert Systems with Applications \textbf{65} (2016)  194--211

\bibitem{Pesic.2006}
Pesic, M., van~der Aalst, W.M.P.:
\newblock A declarative approach for flexible business processes management.
\newblock In Eder, J., Dustdar, S., eds.: Business Process Management
  Workshops, Springer Berlin Heidelberg (2006)  169--180

\bibitem{Samwald.2012}
Samwald, M., Fehre, K., de~Bruin, J., Adlassnig, K.P.:
\newblock The arden syntax standard for clinical decision support: experiences
  and directions.
\newblock Journal of biomedical informatics \textbf{45}(4) (2012)  711--718

\bibitem{Anand.2018}
Anand, V., Carroll, A.E., Biondich, P.G., Dugan, T.M., Downs, S.M.:
\newblock Pediatric decision support using adapted arden syntax.
\newblock Artificial Intelligence in Medicine \textbf{92} (2018)  15--23
  Special Issue on Arden Syntax.

\bibitem{Schuh.2018}
Schuh, C., {de Bruin}, J.S., Seeling, W.:
\newblock Clinical decision support systems at the vienna general hospital
  using arden syntax: Design, implementation, and integration.
\newblock Artificial Intelligence in Medicine \textbf{92} (2018) Special Issue
  on Arden Syntax.

\bibitem{Adlassnig.2015}
Adlassnig, K.P., Fehre, K., Rappelsberger, A.:
\newblock Fuzzy-arden-syntax-based, vendor-agnostic, scalable clinical decision
  support and monitoring platform.
\newblock Studies in health technology and informatics \textbf{216} (2015)
  1111

\bibitem{Caron.2013}
Caron, F.:
\newblock Business process analytics for enterprise risk management and
  auditing.
\newblock PhD thesis, Katholieke Universiteit Leuven, Belgium (2013)

\bibitem{Dunzer.2019}
Dunzer, S., Stierle, M., Matzner, M., Baier, S.:
\newblock Conformance checking: a state-of-the-art literature review.
\newblock In Betz, S., ed.: Proceedings of the 11th International Conference on
  Subject-Oriented Business Process Management, New York, {Association for
  Computing Machinery} (2019)  1--10

\bibitem{Adriansyah.2013}
Adriansyah, A., Munoz-Gama, J., Carmona, J., van Dongen, B.F., van~der Aalst,
  W.M.P.:
\newblock Alignment based precision checking.
\newblock In La~Rosa, M., Soffer, P., eds.: Business Process Management
  Workshops. Volume 132., Springer Berlin Heidelberg (2013)  137--149

\bibitem{Leoni.2013}
de~Leoni, M., van~der Aalst, W.M.P.:
\newblock Aligning event logs and process models for multi-perspective
  conformance checking: An approach based on integer linear programming.
\newblock In Daniel, F., Wang, J., Weber, B., eds.: Business process
  management. Volume 8094 of LNCS sublibrary: SL 3 - Information systems and
  application, incl. Internet/Web and HCI.
\newblock Springer, Heidelberg (2013)  113--129

\bibitem{MannhardtDiss.2018}
Mannhardt, F.:
\newblock Multi-perspective process mining.
\newblock PhD thesis, Technische Universiteit Eindhoven (2018)

\bibitem{Arkad.1991}
Arkad, K., Gill, H., Ludwigs, U., Shahsavar, N., Gao, X.M., Wigertz, O.:
\newblock Medical logic module (mlm) representation of knowledge in a
  ventilator treatment advisory system.
\newblock International journal of clinical monitoring and computing
  \textbf{8}(1) (1991)

\bibitem{Hripcsak.1994}
Hripcsak, G.:
\newblock Writing arden syntax medical logic modules.
\newblock Computers in Biology and Medicine \textbf{24}(5) (1994)  331--363

\bibitem{Vetterlein.2010}
Vetterlein, T., Mandl, H., Adlassnig, K.P.:
\newblock Fuzzy arden syntax: A fuzzy programming language for medicine.
\newblock Artificial Intelligence in Medicine \textbf{49}(1) (2010)  1--10

\bibitem{ImplementationGuides}
Jenders, R.A., Haug, P., Adlassning, K.P.:
\newblock {HL7 Arden Syntax: Implementation Guide}.
\newblock Technical report, Health Level Seven International, Ann Arbor, USA
  (2019)

\bibitem{Patel.1998}
Patel, V.L., Allen, V.G., Arocha, J.F., Shortliffe, E.H.:
\newblock Representing clinical guidelines in glif: individual and
  collaborative expertise.
\newblock JAMIA \textbf{5}(5) (1998)  467--483

\bibitem{Gruger.2022b}
Gr{\"u}ger, J., Geyer, T., Bergmann, R., Braun, S.A.:
\newblock {CGK4PM}: Towards a methodology for the systematic generation of
  clinical guideline process models and the utilization of conformance
  checking.
\newblock BioMedInformatics \textbf{2}(3) (2022)  359--374

\bibitem{XES}
IEEE:
\newblock {Standard for eXtensible Event Stream (XES) for Achieving
  Interoperability in Event Logs and Event Streams}.
\newblock IEEE Std 1849-2016 (2016)  1--50

\end{thebibliography}
\bibliographystyle{splncs}

%
\end{document}